\documentclass[10pt,twocolumn,letterpaper]{article}

\usepackage{cvpr}      %

\usepackage[dvipsnames]{xcolor}
\usepackage[accsupp]{axessibility}  %
\usepackage{times}

\usepackage{graphicx}
\usepackage{amsmath}
\usepackage{amssymb}
\usepackage{soul}
\newcommand{\mathcolorbox}[2]{\colorbox{#1}{$\textstyle #2$}}
\usepackage[symbol]{footmisc}

\def\paperID{87} %
\def\confName{3DV\xspace}
\def\confYear{2024\xspace}

\usepackage{overpic}
\usepackage{enumitem} %
\usepackage{overpic} %
\usepackage{color}
\usepackage{booktabs}

\usepackage{amsfonts}       %
\usepackage{nicefrac}       %
\usepackage{microtype}      %
\usepackage{comment}
\usepackage{mathalpha}
\usepackage{float}
\usepackage{cuted}

\usepackage{multirow}

\usepackage{tabularx}
\usepackage{mathtools}
\usepackage{booktabs}
\usepackage{enumitem}%
\usepackage{verbatim}
\usepackage{subcaption}
\usepackage{amsfonts} %
\DeclareMathSymbol{\shortminus}{\mathbin}{AMSa}{"39}
\DeclareFontFamily{OT1}{pzc}{}
\DeclareFontShape{OT1}{pzc}{m}{it}{<-> s * [1.150] pzcmi8t}{}
\DeclareMathAlphabet{\mathpzc}{OT1}{pzc}{m}{it}
\usepackage[mathscr]{eucal}
\usepackage{comment}
\usepackage{soul}
\usepackage{xfakebold}
\usepackage[font=normalfont,labelfont=bf]{caption}

\newcommand{\fbseries}{\unskip\setBold\aftergroup\unsetBold\aftergroup\ignorespaces}
\makeatletter
\newcommand{\setBoldness}[1]{\def\fake@bold{#1}}
\makeatother

\definecolor{cadmiumgreen}{rgb}{0.0, 0.42, 0.24}

\DeclareMathSymbol{\shortminus}{\mathbin}{AMSa}{"39}

\newcommand{\fig}[1]{Fig~\ref{fig:#1}}
\newcommand{\sect}[1]{Sect~\ref{sect:#1}}

\newcommand{\tab}[1]{Table~\ref{tab:#1}}

\newcommand{\eq}[1]{Eq. (\ref{eq:#1})}

\newcommand{\MethodName}{CoDyNeRF\xspace}

\newcommand{\xbf}{\mathbf{x}}
\newcommand{\xbfcan}{\mathbf{x}_{\text{can}}}

\newcommand{\gammabf}{\pmb{\gamma}}

\newcommand{\rhobf}{\mathbf{\rho}}
\newcommand{\nbf}{\textbf{n}}
\newcommand{\omegabf}{\mathbf{\omega}}

\definecolor{turquoise}{cmyk}{0.65,0,0.1,0.3}
\definecolor{purple}{rgb}{0.65,0,0.65}
\definecolor{dark_green}{rgb}{0, 0.5, 0}
\definecolor{orange}{rgb}{0.8, 0.6, 0.2}
\definecolor{red}{rgb}{0.8, 0.2, 0.2}
\definecolor{darkred}{rgb}{0.6, 0.1, 0.05}
\definecolor{blueish}{rgb}{0.0, 0.3, .6}
\definecolor{light_gray}{rgb}{0.7, 0.7, .7}
\definecolor{greyblue}{rgb}{0.25, 0.25, 1}
\definecolor{drivevid}{RGB}{75,154,250}
\definecolor{meshrig}{RGB}{241,135,40}
\definecolor{nvs}{RGB}{236,71,250}

\usepackage{blindtext}

\renewcommand{\paragraph}[1]{\vspace{1em}\noindent\textbf{#1}.}
\usepackage{environ}
\NewEnviron{smequation}{%
    \begin{equation}
    \scalebox{0.85}{$\BODY$}
    \end{equation}
}
\definecolor{cvprblue}{rgb}{0.21,0.49,0.74}
\usepackage[pagebackref,breaklinks,colorlinks,citecolor=cvprblue]{hyperref}
\hypersetup{colorlinks,linkcolor={blue},urlcolor={red}}  
\usepackage[capitalize]{cleveref}
\crefname{section}{Sec.}{Secs.}
\Crefname{section}{Section}{Sections}
\Crefname{table}{Table}{Tables}
\crefname{table}{Tab.}{Tabs.}

\begin{document}
\title{Controllable Dynamic Appearance for Neural 3D Portraits}

\author{
 ShahRukh Athar\\
  Stony Brook University \\
  \texttt{sathar@cs.stonybrook.edu}
  \and
  Zhixin Shu \\
  Adobe Research \\
  \texttt{zshu@adobe.com} \\
  \and
  Zexiang Xu \\
  Adobe Research \\
  \texttt{zexu@adobe.com} \\
  \and
  Fujun Luan \\
  Adobe Research \\
  \texttt{fluan@adobe.com} \\
  \and
  Sai Bi \\
  Adobe Research \\
  \texttt{sbi@adobe.com} \\
  \and
  Kalyan Sunkavalli \\
  Adobe Research \\
  \texttt{sunkaval@adobe.com} \\
  \and
  Dimitris Samaras \\
  Stony Brook University\\
  \texttt{samaras@cs.stonybrook.edu} \\
}

\maketitle
\
\begin{abstract}
Recent advances in Neural Radiance Fields (NeRFs) have made it possible to reconstruct and reanimate dynamic portrait scenes with control over head-pose, facial expressions and viewing direction. However, training such models assumes photometric consistency over the deformed region e.g. the face must be evenly lit as it deforms with changing head-pose and facial expression. Such photometric consistency across frames of a video is hard to maintain, even in studio environments,
thus making the created reanimatable neural portraits prone to artifacts during reanimation. In this work, we propose \MethodName, a system that enables the creation of fully controllable 3D portraits in real-world capture conditions. \MethodName learns to approximate illumination dependent effects via a dynamic appearance model in the canonical space that is conditioned on predicted surface normals and the facial expressions and head-pose deformations. The surface normals prediction is guided using 3DMM normals that act as a coarse prior for the normals of the human head, where direct prediction of normals is hard due to rigid and non-rigid deformations induced by head-pose and facial expression changes. 
Using only a smartphone-captured short video of a subject for training, we demonstrate the effectiveness of our method on free view synthesis of a portrait scene with explicit head pose and expression controls, and realistic lighting effects. \href{http://shahrukhathar.github.io/2023/08/22/CoDyNeRF.html}{The project page can be found here}.

\end{abstract}
\vspace{-0.2cm}
\section{Introduction}
\label{sec:intro}
The creation of photo-realistic human portraits with explicit control of head-pose and facial expressions remains a topic of active research in the computer graphics and computer vision communities. Fully controllable Neural 3D portraits are necessary in AR/VR applications where an immersive 3D experience is important. Recent advances in neural rendering and novel view synthesis~\cite{nerf,nerfies,NerFACE, SRF, NeuralSceneFlow,DNeRF,  kaizhang2020,  Gao-portraitnerf,Liu-2020-NSV, Zhang-2020-NAA,Bemana-2020-XIN,Martin-2020-NIT,xian2020space} 
have demonstrated impressive image-based rendering of complex scenes and objects. Recently, these methods have also been extended to model and reanimate the human head as shown in \cite{rignerf, nha, imavatar}. These works typically employ a learnable deformation to map the deforming head to a canonical space, where the texture and geometry is predicted and rendered. The canonical space represents a mostly static appearance of the face, akin to a UV texture map, with an optional dependence on expressions \cite{rignerf, nha} in order to capture texture changes induced by them. The deformation, geometry, and texture are learnt via back-propagation using the photometric error with respect to the ground truth. 
In order for such a setup to successfully learn the appearance, deformation, and geometry, the training data must be photometrically consistent. More specifically, the color of a particular position on the human head must remain constant once mapped to the canonical space, regardless of the articulated head-pose and facial expression. However, in realistic lighting conditions, this is rarely the case.

In the real world, there is self-shadowing of the face, and the head casts its shadow on other parts of the scene as it rotates and facial expressions change. Similarly, spatially varying skin reflectance induces specularities that change with viewing angles, head poses, and facial expressions. The assumption of a static appearance in the canonical space is no longer true. As a result, deformable NeRFs trained on such data with a \textit{static canonical appearance} assumption suffer from registration error that leads to blurriness in the renderings and inaccurate reproduction of specularities, shading and shadows. Further, if these models use a canonical space dependent on expression parameters, such as \cite{rignerf, nha}, the aforementioned illumination-dependent effects become entangled with them. This entanglement leads to artifacts in articulated expression and inaccurate reproduction of the illumination effects during reanimation.

Accurately reproducing illumination-dependent effects due to strongly non-ambient lighting requires a modification of the canonical space. It cannot be static, it must be dynamic i.e. the canonical space must vary in appearance as illumination effects change in the deformed space. More specifically, in a dynamic portrait scene that is captured under constant but unknown non-ambient lighting, the following lighting effects must be reproduced in the canonical space  1) Specularities, which depend on viewing directions and surface normals %
2) Shading, that is dependent on head-pose and facial expression deformations as they determine the relative orientation of the surface normals to the lighting and 3) Cast shadows, which are dependent on  whether or not a strong light source is occluded w.r.t a point.
 Capturing each of these effects, especially the cast shadows, using a physically based model is untenable in a deforming NeRF framework due to the exponentially increasing MLP evaluations that need to be performed. Our key insight is that, given enough training data, the implicit and explicit dependencies of the aforementioned illumination effects on the surface geometry, the head-pose, and the facial expression deformations can be approximated by an appropriately conditioned MLP. Based on this, we present \MethodName, a method that uses a dynamic canonical appearance space, modelled by an MLP,  to enable the creation of reanimatable and photorealistic neural 3D portraits using data captured in realistic lighting conditions.  %
\MethodName predicts illumination dependent effects directly in the canonical space by conditioning an MLP on the dynamic surface normals and 3DMM keypoints as well as expression and pose deformations. This conditioning makes it easier for the MLP to interpolate illumination dependent effects for novel facial expressions, head-poses and views without sacrificing the quality of facial expression and head-pose articulation.\\
\indent However, a challenge remains. While 3DMM keypoints, facial expression and head-pose deformations are given by the 3DMM, the surface normals are not. Due to the lack of available ground truth geometry, it is hard to estimate accurate surface normals for each point in the scene, especially on the head and face, which are dynamic and undergo strong deformations with changing facial expression and head-pose. One possible solution is to use the normals given by the density field of the NeRF. However, as shown in \fig{normals_results} and observed in prior work \cite{refnerf}, these are often noisy and inaccurate. Instead, \MethodName uses a carefully designed MLP, that is able to leverage both 3DMM and scene normal priors, to predict the surface normals. The MLP-predicted normals capture more detail than 3DMM-based normals and, due to the 3DMM-prior, transform correctly as the head undergoes rigid and non-rigid deformations. These normals are then used to supervise the gradient density normals of the NeRF in order to ensure accurate reconstruction of the dynamic head geometry and, consequently, the accurate reproduction of illumination effects. Once trained, \MethodName realistically reproduces shadowing, shading and specular effects during reanimation with explicit control of head pose, facial expression and camera viewpoint.\\
\indent In summary, our contributions in this paper are as follows: 
1) Using a dynamic canonical appearance, we are capable of creating a fully reanimatable 3D neural portrait from data captured in strong, non-ambient lighting conditions. \\
2) We propose a method to predict accurate and detailed surface normals of the deforming human head, in addition to the static scene, which is critical for the dynamic appearance learning. \\
3) We enable a realistic re-animation of lighting and specularity effects on the human face with changing head-pose and facial expressions. \\

\begin{figure*}[h]
    \centering
    \includegraphics[width=0.90\linewidth]{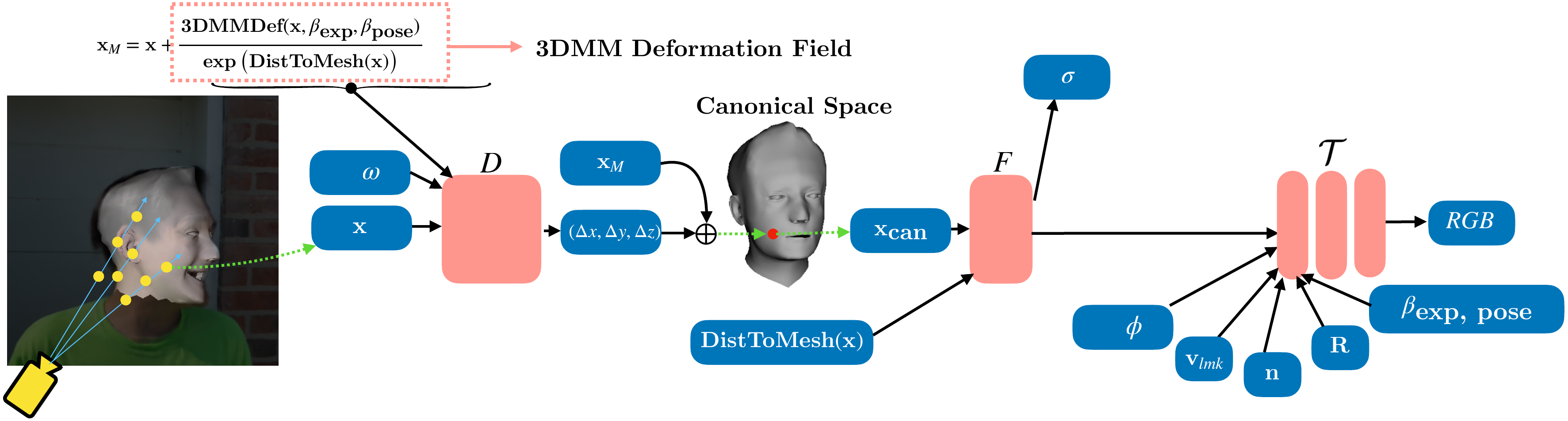}
    
    \caption{\small{\textbf{Overview of \MethodName.} \MethodName is a deformable NeRF architecture that consists of three learnable MLPs: a deformation MLP \(D\) and density MLP \(F\) and a dynamic appearance MLP \(\mathcal{T}\). Given an image, we shoot rays through each of its pixels.
    For every ray, we deform each point on it according to a 3DMM-guided deformation field similar to prior work \cite{rignerf}. Next, the deformed point is given as input to the color MLP, \(F\), which predicts the density and neural features that are passed onto the dynamic appearance MLP \(\mathcal{T}\). \(\mathcal{T}\) then takes as input normals, the reflection vector about the normal, the pose and expression deformations along with spherical harmonics shading and head landmark positions to predict dynamic RGB of the point. The final color of the pixel is calculated via volume rendering.
    }}
    \label{fig:method}
\end{figure*}
\vspace{-0.3cm}
\vspace{-0.3cm}
\section{Related works}
\MethodName  enables  the realistic rendering of illumination effects for controllable neural portraits captured in challenging lighting conditions. It is closely related to recent work on neural rendering, novel view synthesis, 3D face modeling, and controllable face generation.
\\
\paragraph{Neural Scene Representations and Novel View Synthesis} \MethodName is related to recent work in neural rendering and novel view synthesis \cite{imavatar, rignerf, nha,nerf,park2021hypernerf,nerfies,NerFACE, IDR, SRF, lassner2020pulsar, lombardi2021mixture, NeuralSceneFlow,DNeRF, SVS, kaizhang2020, unisurf, Gao-portraitnerf,sitzmann2019scene,Liu-2020-NSV, Zhang-2020-NAA,Bemana-2020-XIN,Martin-2020-NIT,xian2020space, Wizadwongsa2021NeX, refnerf}. Neural Radiance Fields (NeRF) use a Multi-Layer Perceptron (MLP), \(F\), to learn a volumetric representation of a scene. For every 3D point on a ray, NeRFs predict an associated color and density which is volume rendered to give the final color. While NeRFs are able to reproduce specular effects, they do so at the cost of geometric fidelity of the scene \cite{refnerf}. RefNeRF \cite{refnerf} extends NeRFs to explicitly handle specularities of static scenes by improving the learnt surface normals, and consequently the scene geometry.  While NeRFs such as RefNeRF are able to generate photo-realistic images for novel view synthesis, it is only designed for a static scene and is unable to represent scene dynamics. Our approach ensures the accurate reproduction of illumination effects when reanimating neural portraits since, it is specifically designed to model illumination effects of a dynamic scene. 

\vspace{-0.3cm}

\paragraph{Dynamic Neural Scene Representations} There has been an active effort to extend NeRFs to dynamic scenes. Most works do so by imposing temporal constraints either explicitly using scene flow \cite{NeuralSceneFlow, li2021neural, DNeRF, xian2020space}  or  implicitly by using a canonical frame \cite{DNeRF, nerfies}. Authors of \cite{park2021hypernerf} build upon \cite{nerfies} and model topologically changing deformations by lifting the canonical space to high dimensions. The deformation fields in these approaches are conditioned on learnt latent codes without specific physical or semantic meaning, and therefore not controllable in an intuitive manner. Further, such models only work with limited illumination change across frames, since learning a common registration becomes significantly difficult without any photometric consistency.

\vspace{-0.3cm}
\paragraph{Controllable Face Generation} 
Generative Adversarial Networks(GANs)\cite{goodfellow2014generative, pix2pix2016, CycleGAN2017, StyleGAN, Karras-2019-ASB, Karras-2020-AAI} have enabled high-quality image generation and have inspired a large collection of work \cite{shu2018deforming, NeuralFace2017, athar2020self, pumarola2020ganimation, StarGAN2018, starganv2, tewari2020stylerig, tewari2020pie, deng2020disentangled, kowalski2020config} capable of face manipulation. However, it is challenging to enable high-quality view synthesis and 3D controls of the portraits as most of these works lack any 3D understanding and are purely image-based. Works such as \cite{Kim-2018-DVP, Doukas2021Head2HeadDF, head2head2020, facedet3d} fix this by using an intermediate 3D face representation, via a 3D Morphable Model, to reanimate face images/videos. While head poses  and facial expressions are modelled with good detail in these models, thanks to the 3DMM, they are often unable to perform novel view synthesis as they focus on face region but neglect the geometry or appearance of the scene.

In a similar vein, NerFACE\cite{NerFACE}, uses NeRF to model a 4D face avatar and allows pose/expression control on the head. However, no view synthesis can be performed on the scene and the subject is assumed to be uniformly lit throughout the capture process. Neural Head Avatars \cite{nha}, IMAvatar \cite{imavatar} and RigNeRF \cite{rignerf}  improve upon the results of NerFACE further by using a 3DMM prior more explicitly. Neural Head Avatars learns a per-vextex pose and expression conditioned deformation for the FLAME mesh along with a detailed texture while IMAvatar \cite{imavatar} learns a per-point FLAME basis, used for registering different head-poses and facial expressions. RigNeRF uses the 3DMM deformation as a prior on its deformation field that maps points from the articulated space to the canonical space. Unlike our method, all  three aforementioned methods require an ambiently lit face throughout the capture process and are unable to render expression and head-pose dependent illumination effects.

\section{\MethodName}

In this section, we describe our method, \MethodName, that enables the creation of reanimatable 3D portrait scenes from videos captured in the real-world with non-ambient lighting conditions. A deformable Neural Radiance Field (NeRF) \cite{nerf}, with a per-point 3DMM guided deformation field models facial expression and head-pose. It maps points from the deformed space to a dynamic canonical space of the model where the volume density and the appearance is predicted. The dynamic canonical space is conditioned on the surface normals, head-pose and facial expression deformations along with other shading and shadowing based cues (Sec. 3.2). The surface normals, 
 defined in the scene world co-ordinates, are dynamic and vary with head pose and facial expression.
\MethodName predicts these normals using an MLP that is trained with 3DMM normals and scene normals as a prior (Sec. 3.2.1). An overview of the full architecture is shown in \fig{method}.
Once trained, \MethodName is not only able to control facial expression and head pose of the subject but is also able to faithfully capture the varying illumination effects, such as specularities and shadows.

\subsection{A 3DMM-guided Deformable Neural Radiance Field}
A neural radiance field (NeRF) is defined as a continuous function \(F:\left(\gammabf_{m}(\mathbf{x}(t_{i})), \gammabf_{n}(\mathbf{d})\right) \rightarrow (\mathbf{c}(\xbf(t_{i}), \mathbf{d}), \sigma(\xbf(t_{i})))\), that, given the position of a point in the scene \(\mathbf{x}(t_{i}) = \mathbf{o} + t_{i}\mathbf{d}\) that lies on a ray originating at \(\mathbf{o}\) with direction \(\mathbf{d}\), outputs the color \(\mathbf{c} = (r,g,b)\) and the density \(\sigma\). \(F\) is usually represented as a multi-layer perceptron (MLP) and \(\gammabf_{m}: \mathbb{R}^{3} \rightarrow \mathbb{R}^{3 + 6m}\) is the positional encoding \cite{nerf} defined as \(\gammabf_{m}(\xbf) = (\xbf,...,\text{sin}(2^{k}\xbf(t_{i})),\text{cos}(2^{k}\xbf(t_{i})),...)\) where \(m\) is the total number of frequency bands and \(k \in \{0,...,m-1\}\). The expected color of the pixel through which a camera ray 
passes is calculated via volume rendering as follows:
\begin{smequation}
\begin{split}
   &C = \sum_{t}\omega_{t}c(x(t)); \\
   &\text{ where } w_i=\text{\footnotesize{exp}}(\shortminus \sum_{j<i} \sigma_j\left(t_{j+1}-t_j\right))\left(1\shortminus \text{\footnotesize{exp}}(\shortminus \sigma_i\left(t_{i+1}\shortminus t_i\right))\right)  
\end{split}
\label{eq:vol_rend}
\end{smequation}
The parameters of \(F\) are trained to minimize the L2 distance between the expected color and the ground-truth.

  NeRFs can be extended to model dynamic scenes by using a deformation field to map each 3D point of the scene to a canonical space, where the volumetric rendering takes place \cite{rignerf, nerfies, DNeRF, park2021hypernerf}. The deformation field is also represented by an MLP \(D_{i}: \xbf \rightarrow \xbfcan\) where  \(D_{i}\) is defined as \(D(\xbf, \omega_{i}) = \xbfcan\) and \(\omega_{i}\) is a per-frame latent deformation code. Following prior work \cite{rignerf}, we use a 3DMM prior on the deformation field as follows:
\begin{smequation}
    \begin{split}
        \hat{D}(\xbf) &= \text{3DMMDef}(\xbf, \beta_{\text{i,exp}}, \beta_{\text{i,pose}}) \\
                       & + D(\gammabf_{a}(\xbf), \gammabf_{b}(\text{3DMMDef}(\xbf, \beta_{\text{i,exp}}, \beta_{\text{i,pose}})), \omega_{i})\\
        \xbfcan =  & \xbf +  \hat{D}(\xbf)
    \end{split}
    \label{eq:rignerfcanmap}
\end{smequation}
where, \(\text{3DMMDef}(\xbf, \beta_{\text{i,exp}}, \beta_{\text{i,pose}})\) is the deformation prior given by the 3DMM, \(\beta_{\text{i,exp}}, \beta_{\text{i,pose}}\) are the articulated facial expression and head-pose of the frame \(i\),  and \(\gammabf_{a}, \gammabf_{b}\) are the positional encoding functions with frequencies \(a \text{ and } b\) respectively. The deformation prior,  can be written as follows:

\begin{smequation}
    \text{3DMMDef}(\xbf, \beta_{\text{i,exp}}, \beta_{\text{i,pose}}) = \frac{\text{3DMMDef}(\hat{\xbf}, \beta_{\text{exp}}, \beta_{\text{pose}})}{\text{exp}(\text{DistToMesh}(\xbf))}
\end{smequation}
where, \(\hat{\xbf}\) is the closest point on the mesh to \(\xbf\), \(\text{DistToMesh}(\xbf) = ||\xbf - \hat{\xbf}||\) is the distance between \(\xbf\) and \(\hat{\xbf}\) and \(\text{3DMMDef}(\hat{\xbf}, \beta_{\text{i,exp}},\beta_{\text{i,pose}})\) is the deformation of the vertex \(\hat{\xbf}\) as follows:

\begin{smequation}
    \text{3DMMDef}(\hat{\xbf},\beta_{\text{exp}}, \beta_{\text{pose}}) = \hat{\xbf}_{\text{FLAME}(\beta_{\text{exp, can}}, \beta_{\text{pose, can}})} - \hat{\xbf}_{\text{FLAME}(\beta_{\text{exp}}, \beta_{\text{pose}})}
\end{smequation}
where, \(\hat{\xbf}_{\text{FLAME}(\beta_{\text{exp, can}}, \beta_{\text{pose, can}})}\) is the position of \(\xbf\) in the canonical space and \(\hat{\xbf}_{\text{FLAME}(\beta_{\text{exp}}, \beta_{\text{pose}})}\) is its position with head pose and facial expression parameters \(\{\beta_{\text{exp}}, \beta_{\text{pose}}\}\).
\subsection{An Illumination aware dynamic canonical appearance model}
\label{sect:dyconnerf}
In the canonical space, \MethodName predicts the density and a dynamic RGB appearance. The dynamic RGB is conditioned on the surface normals, head-pose, and expression deformations along with other shading and shadowing cues such, as the reflection vector and global location of the head. Below, we describe each aspect of the appearance model. %

\vspace{-0.2cm}

\paragraph{A spatially conditioned density prediction model} First, \MethodName predicts the density at any point by conditioning on its position in the canonical space and its distance to the mesh:

\begin{smequation}
    \sigma(\xbf), \tau = F(\gammabf_{c}(\xbfcan), \text{DistToMesh}(\xbf))
        \label{eq:rig_nerf_density}
\end{smequation}
where, \(F\) is an MLP, \(\tau\) is a feature vector, \(\text{DistToMesh}(\xbf) = ||\xbf - \hat{\xbf}||\) is the distance of \(\xbf\) to the closest mesh vertex \(\hat{\xbf}\) and \(\gammabf_{c}\) is the positional encoding function with \(c\) frequencies. Additional conditioning on \(\text{DistToMesh}(\xbf)\) modestly boosts both the training speed and quality of results by allowing \(F\) to distinguish between points in the canonical space that have never been deformed and points that have been deformed to the canonical space. We provide experiments supporting this in the supplementary section.

\textbf{An illumination aware dynamic canonical appearance model.} Next, \MethodName predicts dynamic RGB conditioned on inputs that capture local geometry, surface properties and viewing direction. 
Since the captured neural portrait is a dynamic scene, the outgoing radiance at any point \(\xbf\), is implicitly dependent on facial expression and head-pose, \(\{\beta_{\text{exp, pose}}\}\) (or \(\{\beta_{\text{e,p}}\}\)) due surface properties and incoming radiance being dependent on them. More specifically, at any point \(\xbf\) for a particular articulation of facial expression and head-pose, \(\beta_{\text{e,p}}\), the outgoing radiance is given by the rendering equation as follows:

\begin{smequation}
    L_r\left(\mathbf{x}, \omegabf_o, \beta_{\text{e,p}}\right)=\int_{\omega_i} \rhobf\left(\mathbf{x}, \omegabf_i, \omegabf_o, \beta_{\text{e,p}}\right)\left(\mathbf{n} \cdot \omegabf_i\right) L_i\left(\mathbf{x}, \omegabf_i, \beta_{\text{e,p}}\right) \mathrm{d} \omega_i
    \label{eq:rend_eq}
\end{smequation}
where, \(\rhobf\) is the articulation dependent BRDF, \(\nbf\) is the normal at \(\xbf\) and \(\omega_i, \omega_o\) are the incoming and outgoing ray directions respectively. We approximate this integral in the canonical space by an MLP, \(\mathcal{T}\), as follows:
\begin{smequation}
    (R,G,B) = \mathcal{T}(\tau, \textbf{n}, \textbf{R},\textbf{v}_{lmk}, \text{3DMMDef}_{\text{exp}},  \text{3DMMDef}_{\text{pose}}, \phi_{i})
    \label{eq:lt_pred}
\end{smequation}
where, \(\tau\) are features from the density prediction network from \eq{rig_nerf_density}, \(\textbf{n}\) is the surface normal, \(\textbf{v}_{lmk}\) are the facial landmarks, \(\textbf{R} = 2(\mathbf{d}.\mathbf{n})\mathbf{n} - \mathbf{d}\) is the reflection vector, \(\text{3DMMDef}_{\text{exp}} \coloneqq \text{3DMMDef}(\xbf, \beta_{\text{exp}}, \beta_{\text{pose, can}})\) is the expression-only deformation given by the 3DMM, \(\text{3DMMDef}_{\text{pose}}\coloneqq \text{3DMMDef}(\xbf, \beta_{\text{exp, can}}, \beta_{\text{pose}})\) is the head-pose only deformation given by the 3DMM and \(\phi_{i}\) is a per-frame latent vector that is learnt through optimization. Each input in \eq{lt_pred} contains information that is essential to the prediction of accurate illumination effects. First, surface reflectance and absorption properties are captured by \(\tau\) which is predicted in the canonical space and thus is forced to only model deformation-independent properties of the surface. The surface normal \(\textbf{n}\) is used to model shading effects and, along with the reflection vector \(\textbf{R}\),  specular effects. The face landmarks, \(\textbf{v}_{lmk}\), along with expression and head-pose deformations,\(\text{3DMMDef}_{\text{exp}} \text{ and } \text{3DMMDef}_{\text{pose}}\) are used to model cast shadows, inter-reflections and any other illumination effects that depend on the global orientation of the head and deformations due to facial expressions and head-pose. The latent-code captures any appearance changes due to the camera. 

\subsubsection{Prediction of Dynamic Surface Normals}

\begin{figure}[h]
    \centering
    \includegraphics[width=0.45\linewidth]{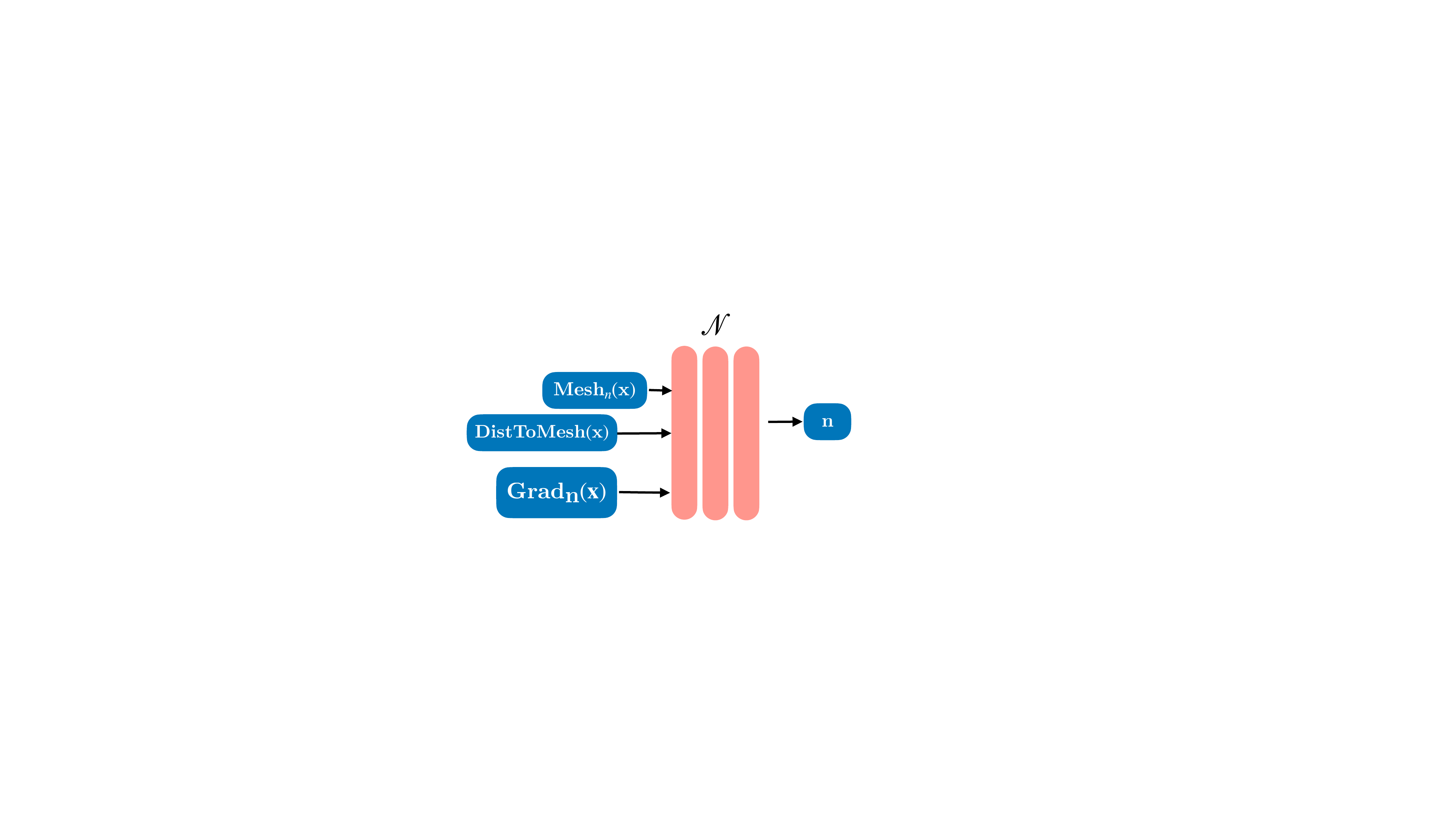}
    
    \caption{\small{\textbf{Normals Prediction Architecture.} The Normals prediction network takes as input the mesh normals of a given point \(\xbf\), its distance to the mesh and the normals given by gradient of the NeRF's density field. The details of the architecture are discussed in Section 3.2.1. 
    }}
    \label{fig:normalsnet}
\end{figure}
\vspace{-0.1cm}
Prediction of shading and specular effects (through  reflection vector \(\mathbf{R}\)) requires accurate surface normals. Within a NeRF, one straightforward way to calculate the normals at any point \(\xbf\) is to define it as the negative of the derivative of the density field, \(\sigma(\xbf)\), w.r.t \(\xbf\) i.e \(\text{Grad}_{\textbf{n}}(\xbf) = -\frac{\nabla_{\xbf} \sigma(\xbf)}{|\nabla_{\xbf} \sigma(\xbf)|}\). 
However, as  shown in \fig{normals_results} and observed in prior work \cite{refnerf}, unless these normals are regularized \cite{refnerf}, they are incredibly noisy and rather unusable.  %
  To get around this, we use an MLP, \(\mathcal{N}\), to predict the normals. We design \(\mathcal{N}\) in a manner that allows it to exploit local priors from the 3D head mesh and the scene to predict the normals at each point of the dynamic neural portrait. More specifically, the normal at any point is given as follows:
\begin{smequation}
    \textbf{n} = \mathcal{N}(\text{Mesh}_{\textbf{n}}(\textbf{x}), \text{Grad}_{\textbf{n}}(\xbf), \text{DistToMesh}(\xbf))
\end{smequation}
where, \(\text{Mesh}_{\textbf{n}}(\textbf{x})\) is normal vector of the mesh vertex closest to \(\xbf\), \(\text{Grad}_{\textbf{n}}(\xbf)\) is the normal calculated by the negative gradient of the density w.r.t the input point and \(\text{DistToMesh}(\xbf)\) is the distance of \(\xbf\) to the mesh. With these three inputs, \(\mathcal{N}\), is able to rely on the 3DMM mesh normals for points close to the head, while relying on gradient normals \(\text{Grad}_{\textbf{n}}(\xbf)\) everywhere else. In supplementary, be demonstrate the utility of each input to \(\mathcal{N}\) by ablating each one of them. 

We train \(\mathcal{N}\) through a combination of weak supervision on mesh and scene normals, and regularization losses. The prediction of \(\mathcal{N}\) is forced to be weakly consistent with the 3DMM on its vertices as follows:
\begin{figure}[t]
    \centering
    \includegraphics[width=0.9\linewidth]{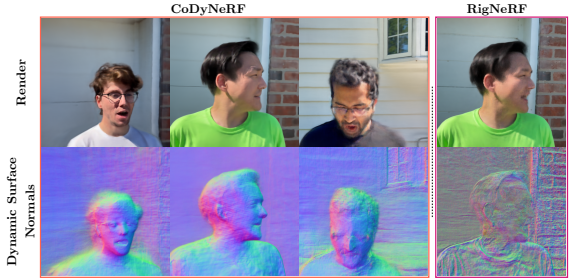}
    
    \caption{\small{\textbf{Dynamic Surface Normals.} Surface normals directly influence the specularities and shading of the human head. Here we visualise the surface normals given by the density field of \MethodName. As can be seen, the normals are able to capture the dynamic foreground with fine detail without sacrificing accuracy. In comparison, the normals of RigNeRF \cite{rignerf} are noisy and fail to capture the facial geometry accurately.
    }}
    \label{fig:normals_results}
\vspace{-0.2cm}
\end{figure}
\begin{smequation}
\mathcal{R}_{Mesh, \textbf{n}} = \lambda_{Mesh,\textbf{n}}\sum_{v}||\mathcal{N}(v) - \text{Mesh}_{\textbf{n}}(v)||
\end{smequation}
where, \(v\) are the vertices of the mesh and \(\lambda_{Mesh,\textbf{n}}\) is the regularization constant. The normals predicted by \(\mathcal{N}\) are also forced to be forward facing by using the following regularization \cite{refnerf}:
\begin{smequation}
\mathcal{R}_{dir, \textbf{n}}=\sum_i w_i(\xbf_{i}) \max \left(0,  \mathcal{N}(\xbf_{i}) \cdot \mathbf{d}_{i}\right)^2
\end{smequation}
where, \(\xbf_{i}\) are points along the ray passing through pixel \(i\) with direction \(\mathbf{d}_{i}\) and \(w_i(\xbf_{i})\) is  the weight of \(\xbf_{i}\) per \eq{vol_rend}. In order to ensure the gradient density normals \(\text{Grad}_{\textbf{n}}(\xbf)\) are themselves accurate, we regularize both the normals predicted by \(\mathcal{N}\) and the gradient density normals to be consistent with each other:
\begin{smequation}
\mathcal{R}_{\textbf{n}} = \sum_i w_i(\xbf_{i}) ||\mathcal{N}(\xbf_{i}) -  \text{Grad}_{\textbf{n}}(\xbf)||
\end{smequation}
As observed in \cite{refnerf}, this regularization ensures that highly specular surfaces of the scene are not explained away as subsurface emmisive lobes.
The full loss on \(\mathcal{N}\) is:
\begin{smequation}
\begin{split}
    \mathcal{L}_{\mathcal{N}} &= \underbrace{\mathcal{R}_{Mesh, \textbf{n}}}_{\substack{\text{Ensures consistency} \\ \text{of normals on vertices of the \textit{head}}}} + \underbrace{\mathcal{R}_{dir, \textbf{n}}}_{\substack{\text{Ensures normals are}\\\text{forward facing}}} +\underbrace{\mathcal{R}_{ \textbf{n}}}_{\substack{\text{Ensures consistency between} \\ \text{\(\mathcal{N}\) and \(\textbf{Grad}_{\textbf{n}}\) normals}}} 
\end{split}
\label{eq:normals_full_loss}
\end{smequation}

\paragraph{Sparsity regularization for fast calculation of \(\mathcal{R}_{dir, \textbf{n}}\) and \(\mathcal{R}_{ \textbf{n}}\)}
Calculating both \(\mathcal{R}_{dir, \textbf{n}}\) and \(\mathcal{R}_{ \textbf{n}}\) is very computationally expensive as it requires a second derivative calculation at each point along the ray (usually \(\sim\) 100 points for most standard NeRF architectures) for each sampled ray in the batch (typically around 1000 rays). A simple way to reduce the computational burden is to evaluate the above sum only on a subset of the points on a ray as follows
\begin{smequation}
\begin{split}
    \mathcal{R}_{\textbf{n}} = &\sum_i w_i(\xbf'_{i}) ||\mathcal{N}(\xbf'_{i}) -  \text{Grad}_{\textbf{n}}(\xbf'_{i})||; \\
    &\text{ where } \xbf'_{i} \in \mathcal{S}_{i,k}
\end{split}
\label{eq:imp_normals_reg}
\end{smequation}

\noindent where, \(\mathcal{S}_{i,k}\) is the set of top \(k\) points, sorted by weight \(w_i(\xbf'_{i})\), of the ray passing through pixel \(i\). However, as the weights predicted by the NeRF are relatively broadly distributed, such regularization does not minimize \eq{imp_normals_reg} over the whole scene consistently. To ensure the predicted weights are more tightly distributed around the surface, we enforce a Cauchy regularization to enforce sparsity \cite{sgnerf}
\begin{smequation}
\mathcal{R}_{cauchy} = \lambda_{c} \sum_{i} \log \left(1+\frac{\sigma\left(\xbf_{i}\right)^2}{c}\right)
\label{eq:cauchy_normals_reg}
\end{smequation}
similar to \cite{sgnerf}, we only apply this on the coarse MLP. \eq{imp_normals_reg} and \eq{cauchy_normals_reg} improve the underlying dynamic scene geometry and significantly improve the quality of the gradient density normals.
\begin{figure*}[h]
    \centering
    \includegraphics[width=0.95\linewidth]{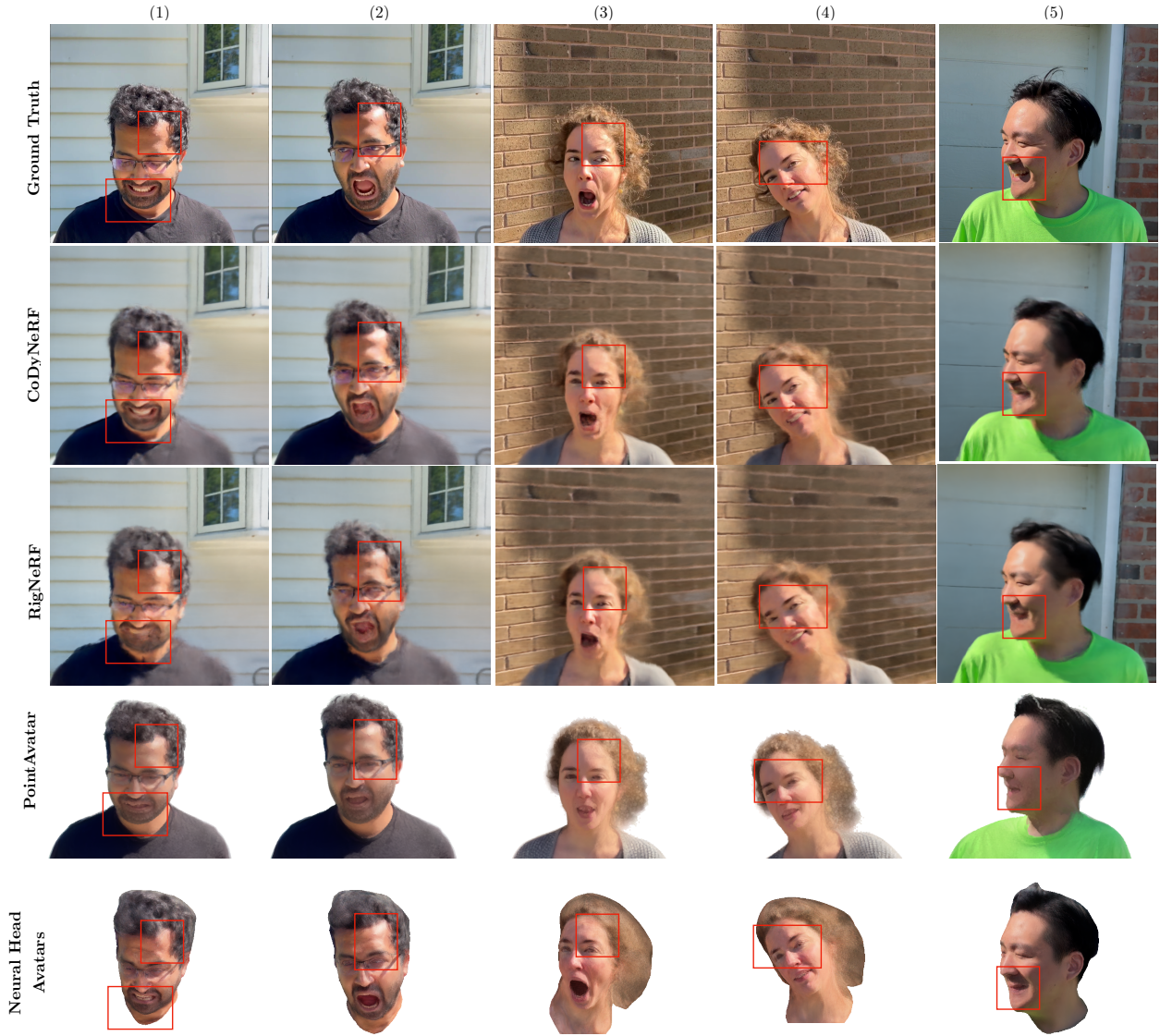}
    
    \caption{\small{\textbf{Qualitative comparison by reanimation with novel facial expression, head-pose, and camera view parameters.} Here we reanimate \MethodName, RigNeRF \cite{rignerf} and Neural Head Avatars \cite{nha} with novel facial expressions and head-pose extracted from the ground-truth frame in Row 1. As can be observed in the highlighted red boxes, while RigNeRF \cite{rignerf} is able to generate some shadowing and specular effects on the face, they are incorrect. Additionally, RigNeRF's results have significant artifacts around the mouth and face regions. Similarly, due to the use of an explicit mesh and the entanglement of illumination effects with expressions, Neural Head Avatars \cite{nha} is unable to recover accurate geometries or predict accurate illuminations effects. Since PointAvatar's design does not take into account cast shadows and specularities, it struggles to reproduce them accurately as can be seen in columns (1) - (5). In contrast, our approach, \MethodName, is able to accurately reproduce cast shadows and specularities without sacrificing the quality of facial expressions and head-pose articulation. 
    }}
    \label{fig:val_results}
    \vspace{-0.2cm}
\end{figure*}

\section{Experimental Results}

\begin{table*}[t]
\begin{center}
\small
\scalebox{0.65}{
\begin{tabular}{lccccccccccccc}
\toprule
  & \multicolumn{3}{c}{\emph{Subject 1}}  &  \multicolumn{3}{c}{\emph{Subject 2}} &  \multicolumn{3}{c}{\emph{Subject 3}} &  \multicolumn{3}{c}{\emph{Subject 4}}\\
  \midrule
  Models   & PSNR $\uparrow$ &LPIPS $\downarrow$ & FaceMSE $\downarrow$  
   & PSNR $\uparrow$ &LPIPS $\downarrow$ & FaceMSE $\downarrow$
  & PSNR $\uparrow$ &LPIPS $\downarrow$ & FaceMSE $\downarrow$
  & PSNR $\uparrow$ &LPIPS $\downarrow$ & FaceMSE $\downarrow$\\
\midrule
 \MethodName (Ours) &
 \mathcolorbox{pink}{23.46} & \mathcolorbox{pink}{0.40} & \mathcolorbox{pink}{1.92e-3} &
 \mathcolorbox{pink}{21.24} & \mathcolorbox{pink}{0.33} & \mathcolorbox{pink}{2.16e-3} &
 \mathcolorbox{pink}{22.3} & \mathcolorbox{pink}{0.53} & \mathcolorbox{pink}{1.0e-3} &
 \mathcolorbox{pink}{23.56} & \mathcolorbox{pink}{0.40} & \mathcolorbox{pink}{1.5e-3}
 \\
 RigNeRF \cite{rignerf} &
 21.0 &  0.41  & 3.64e-3  & 
 20.86 &  0.35  & 3.60e-3 &
 21.67 &  0.54  & 1.7e-3 &
 22.26 &  0.45  & 2.1e-3
  \\
 NHA \cite{nha} &
  - &  - & 0.058  & 
  - &  -  & 0.0382 &
  - &  -  & 0.070 &
  - &  -  & 0.068
\\ 
PointAvatar \cite{pointavatar} &
  - &  -  & 6.3e-3 & 
  - &  -  & 8.38e-3 &
  - &  -  & 1.05e-2 &
  - &  -  & 7.74e-3
 \\ 
\bottomrule
\end{tabular}}
\caption{\small{Quantitative results of Subject 1,2,3 and 4 on test data. Here we calculate PSNR and LPIPS over the full image while FaceMSE is only restricted to the MSE calculated over the face region. Our results are better than RigNeRF \cite{rignerf}, Neural Head Avatars \cite{nha} and IMAvatar \cite{imavatar} across all subjects.}
}

\label{tab:Subjects_metrics}
\vspace{-0.3cm}
\end{center}
\end{table*}
\vspace{-0.3cm}

\begin{figure*}[h]
    \centering
    \includegraphics[width=1.0\linewidth]{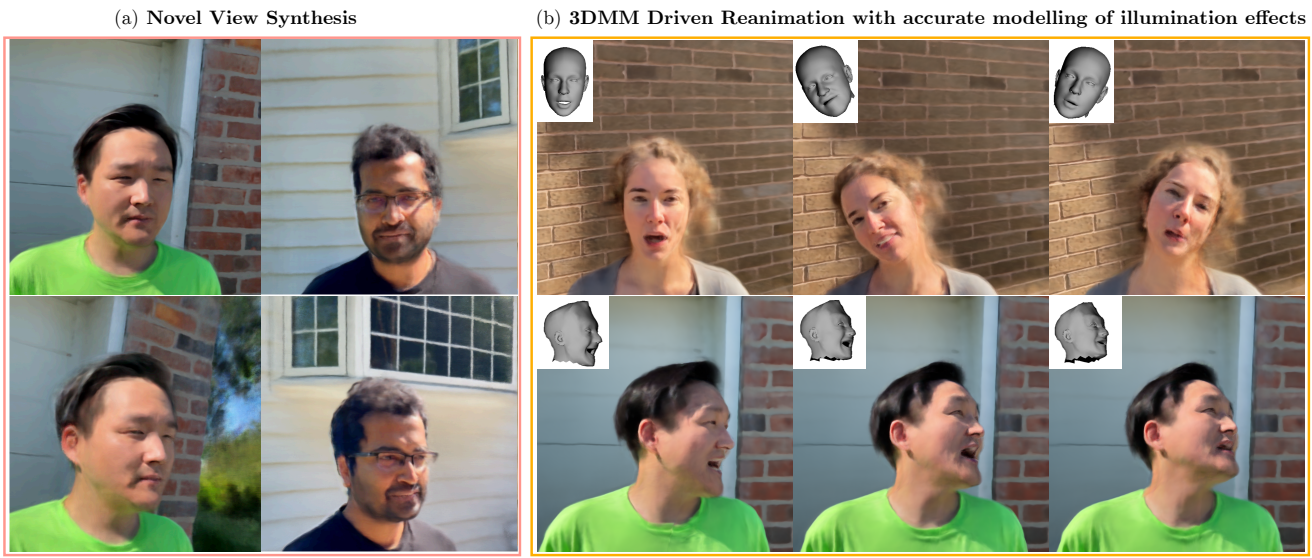}
    \caption{\small{Applications of \textbf{\MethodName}. \MethodName enables the creation of Neural 3D Portraits captured in challenging lighting conditions with the accurate reproduction of illumination effects during novel expression and pose reanimation. In (a), we demonstrate that specularities, especially on the forehead and the nose, change realistically with changing viewing directions. In (b) we  demonstrate the realistic reproduction of shading, shadows and specularities when the neural portrait is reanimated with a driving 3DMM (shown as white insets in the top left of each image). We see from results in the top row of (b) that the shading on the forehead and the specularities change realistically. Similarly, in the bottom row of (b) we see the nose being unshaded as the head rotates and the shadow cast by the ear changes realistically. 
    }}
    
    \label{fig:all_control}
\end{figure*}

\begin{table}
\begin{center}
\small
\scalebox{0.50}{
\begin{tabular}{lcccccccc}
\toprule
  & \multicolumn{3}{c}{\emph{Subject 1}}  &  \multicolumn{3}{c}{\emph{Subject 2}}  \\
  \midrule
  Models   & PSNR $\uparrow$ &LPIPS $\downarrow$ & FaceMSE $\downarrow$  
   & PSNR $\uparrow$ &LPIPS $\downarrow$ & FaceMSE $\downarrow$
  \\
\midrule
 \MethodName \(-\beta_{exp, pose}\), \(-normals\),\(-v_{lmk}\) &
 19.8 & 0.47 & 4.03e-3 &
 19.1 & 0.51 & 4.26e-3 &
 
 \\
 \MethodName  \(-v_{lmk}\) &
 22.1 &  0.45  & 2.64e-3  & 
 20.93 & 0.33 & 3.30e-3 &
 
  \\
 \MethodName (Full Model)  &
  \mathcolorbox{pink}{23.46} & \mathcolorbox{pink}{0.40} & \mathcolorbox{pink}{1.92e-3} &
 \mathcolorbox{pink}{21.24} & \mathcolorbox{pink}{0.37} & \mathcolorbox{pink}{2.46e-3} \\

\bottomrule
\end{tabular}}
\vspace{-0.3cm}
\caption{\small{Ablation of dynamic appearance conditioning.} %
}
\vspace{-0.6cm}
\label{tab:3dmm_aba}
\vspace{-0.3cm}
\end{center}
\end{table}

In this section, we show results of head-pose control, facial expression control, and novel view synthesis using \MethodName. For each scene, the model is trained on a short portrait video captured using a consumer smartphone.

\paragraph{Baseline approaches}  We compare \MethodName quantitatively and qualitatively to the following prior work for human head animation 1) RigNeRF \cite{rignerf} is a method for reanimating neural portraits with full control of facial expressions, head pose and viewing direction. Similar to us, the authors use a volumetric representation to model the dynamic scene. 2) Neural Head Avatars \cite{nha} creates a neural head model by deforming a mesh with pose and expression dependent offsets along with an MLP dependent texture space. 3) PointAvatar \cite{pointavatar} uses a point cloud and an SDF as the geometric representation of the avatar in the canonical space. Additionally, PointAvatar seperates the appearnace of the avatar into an albedo and a normals-conditioned RGB shading. RigNeRF \cite{rignerf} and Neural Head Avatars \cite{nha} use an expression and pose dependent canonical space while PointAvatar uses a normals conditioned shading network to handle apperance variation during the capture. Unlike in \cite{rignerf}, we \textit{do not optimize} the appearance or deformation latent code during testing. Since we want to evaluate the fidelity of reproduction of illumination dependent effects for novel head-pose, facial expressions and views we cannot assume access to testing frames. We always use the deformation and appearance code of the first frame during testing.  %

\paragraph{Training and evaluation Data} The training and validation data was captured using an iPhone 13 Pro Max for all the experiments in the paper. In the first half of the capture, we ask the subject to enact a wide range of expressions and speech while trying to keep their head still as the camera is panned around them. In the next half, the camera is fixed at head-level and the subject is asked to rotate their head as they enact a wide range of expressions. Camera parameters are calculated using COLMAP \cite{Schonberger-2016-SFM}. FLAME  \cite{FLAME:SiggraphAsia2017} parameters and spherical harmonics coefficients \(L_{lm}\)  are obtained via standard photometric and landmark fitting obtained by \cite{3DDFA_V2}. All videos are between 50-80 seconds long; we use the first \(\sim\) 1200-1500 frames for training and the remaining 120-150 \textit{held out} frames, with novel expressions and head-poses, for validation. Please find full details of each experiment in the supplementary document. %

\vspace{-0.1cm}
\subsection{Evaluation on Test Data}
\label{sect:eval_data}
We evaluate \MethodName, along with three state-of-the-art baselines, Neural Head Avatars \cite{nha}, RigNeRF \cite{rignerf}, and PointAvatar \cite{pointavatar} on held out testing frames. These frames contain a variety of facial expressions and head-poses. In \fig{val_results}, we show a qualitative comparison between \MethodName and the baselines.  
We observe that outdoor-captured videos, where pose and expression deformations under sunlight creates large appearance variations, pose a significant challenge to existing methods\cite{pointavatar, nha, rignerf}.
Neural Head Avatars \cite{nha} and RigNeRF \cite{rignerf}, entangle illumination dependent effects with expression parameters in their canonical space and are not able to faithfully reconstruct the target appearance, creating artifacts in the overall appearance including in shadows, specularities, and in the mouth and eye region. 
Neural Head Avatars \cite{nha} is able to generate some shadows and specularities but they often lack of details and do not match the ground truth well. 
Further, due to their heavy reliance on an explicit face mesh, results from \cite{nha} are unable to accurately capture the head geometry in detail, resulting in an unnaturally deformed shape. 
RigNeRF \cite{rignerf}, similar to NHA \cite{nha}, uses an expression and pose (through deformation features) conditioned canonical space and is able to generate some shadows and specularities. However, they are often inaccurate and visually unnatural. For example, the specularity columns 1 and 2 of \fig{val_results} are incorrect, as well as the specularity around the eyes of columns 3 and 4. Similarly, the shadow on the mouth of column 5 is incorrect. Additionally, RigNeRF \cite{rignerf} results have significant artifacts, especially around the mouth and the eyes. artifacts around the mouth can be seen in columns (1), (2), (3) and (5) and there are artifacts around the eyes in columns (2), (3) and (4) of \fig{val_results} respectively. PointAvatar \cite{pointavatar} uses separate shading and texture networks to disentangle the albedo from the shading. However, its design does not take into account cast shadows and specularities, thus it is unable to learn and predict them\footnote{We analyse this further in the appendix section.}. PointAvatar is unable to reproduce the shadows and specularities on the forehead in columns (1) - (4) of \fig{val_results}. It is also unable to predict the shadow cast by the nose on the mouth in column (5) of \fig{val_results}. In contrast, \MethodName can faithfully reproduce the shadows and specularities on the forehead in columns (1) - (4) and the shadow cast by the nose on the mouth in column (5) of \fig{val_results}.

In contrast to prior work, \MethodName is able to faithfully reproduce illumination effects and generate high quality renders. As can be seen in \fig{val_results}, our method captures dynamic appearance details, such as shadow patterns and specularities, under varying pose and expression more accurately \textit{without} sacrificing the quality of expression and head-pose articulation. 
In \tab{Subjects_metrics}, we provide quantitative evaluation of these four methods. We report image similarity measurement (PSNR), perceptual quality (LPIPS) and quality of facial appearance reconstruction (FaceMSE i.e MSE measured over the face region only)  on novel expressions and head-poses. As we can see, \MethodName outperforms previous approaches across all subjects.
\subsection{Ablation of the Dynamic Canonical apperance Model}
In \tab{3dmm_aba}, we have ablated the conditioning of the canonical appearance model on 3DMM-based inputs and normals. First, we measured the performance of model with a static canonical space. This model does not use any surface normals or 3DMM-based information as input, and as can be seen in \tab{3dmm_aba}, it performs quite poorly.  When the surface normals, expression and pose parameters are added, the results are much better, but the model still lacks global head-position conditioning via the landmarks i.e \(v_{lmk}\), and thus is unable to reason about self-shadowing of the face. Our full model (third row), with conditioning from the surface normals, 3DMM expression, pose and the landmarks performs the best.

\section{Limitations and Conclusion}

\MethodName has certain limitations that we address in this section. Similar to prior work, \MethodName is subject specific and trains an individual model per scene. It also requires training sequences of atleast 40 seconds in order to sample the expression and head-pose space sufficiently. \MethodName also does not support relighting with novel lighting. %
Being a method that allows photorealistic facial reanimation, \MethodName may have potentially negative societal impact if misused. We discuss this further in the supplementary.

In conclusion, we present \MethodName, a method that enables the creation of reanimatable Neural 3D Portraits captured in challenging lighting conditions.
Using a dynamic canonical appearance representation, \MethodName is able to accurately reproduce illumination effects with varying facial expression and head-pose. Additionally, we also propose a dynamic normals prediction module that utilizes 3DMM priors, along with a importance sampling based regularizer, to predict accurate dynamic surface normals. We believe this work takes us a step closer to creating in-the-wild Neural 3D portraits that are captured with casual smartphone devices.

\section{Acknowledgements}
This work was supported in part by the CDC/NIOSH through grant U01 OH012476 and a gift from Adobe.

{
    \clearpage
    \small
    \bibliographystyle{ieee_fullname}
    \bibliography{macros,main}
}

\clearpage

\begin{strip}
\centering{\textbf{ \huge Controllable Dynamic Appearance for Neural 3D Portraits}\\~\\ \textbf{\huge -Appendix-} }
\end{strip}

\maketitle
{
  \hypersetup{linkcolor=black}
  \tableofcontents
}

\section{Comparisons against IMAvatar}
\begin{figure}[h]
    \centering
    \includegraphics[width=0.9\linewidth]{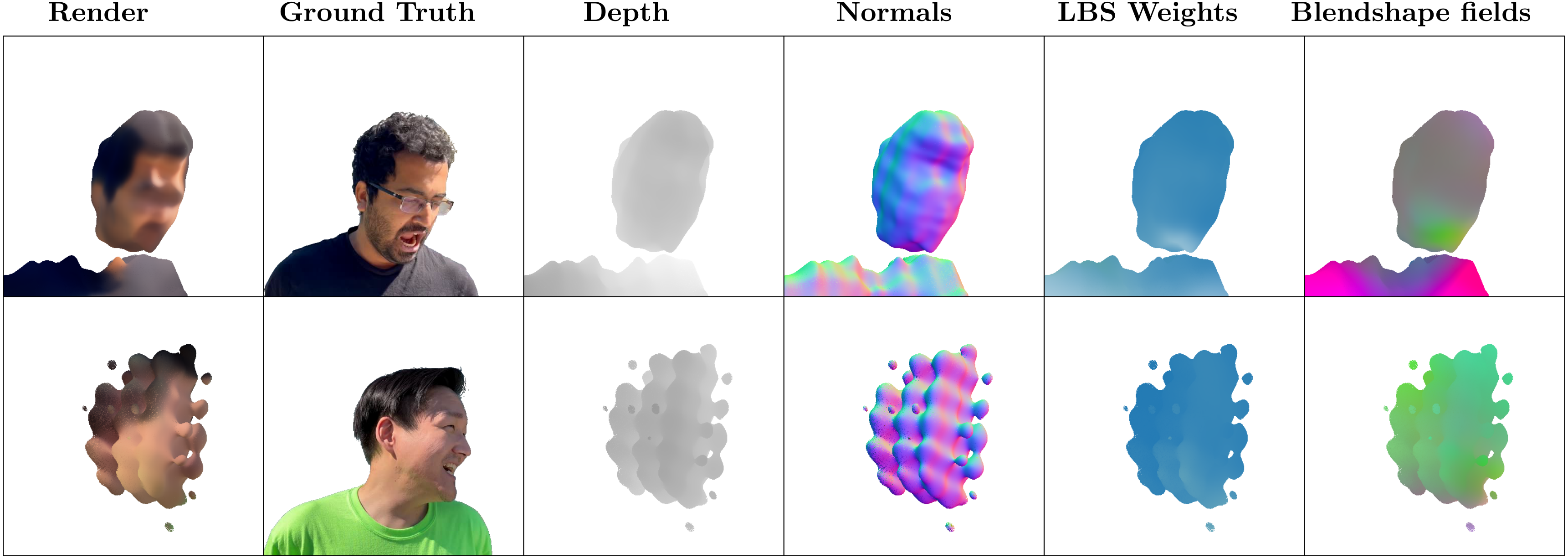}
    \caption{Results of IMAvatar \cite{imavatar} on our dataset. Due to the changing illumination effects induced by head-pose variation in strong non-ambient lighting, IMavatar fails to converge.}
    \label{fig:IMavatar-fail-real}
\end{figure}

In this section we provide details about comparisons against IMAvatar\footnote{We use the code from their official repo at: https://github.com/zhengyuf/IMavatar}. When IMavatar is trained on our dataset, which is captured under strong-non ambient lighting, we find that it is unable to converge to a reasonable result (see \fig{IMavatar-fail-real}). In order to isolate the strong non-ambient lighting as the reason for failure we use synthetic data. 
\subsection{Ambient and Non-Ambient Synthetic Dataset}
\label{sect:syn-data}
\begin{figure}[h]
    \centering
    \includegraphics[width=0.9\linewidth]{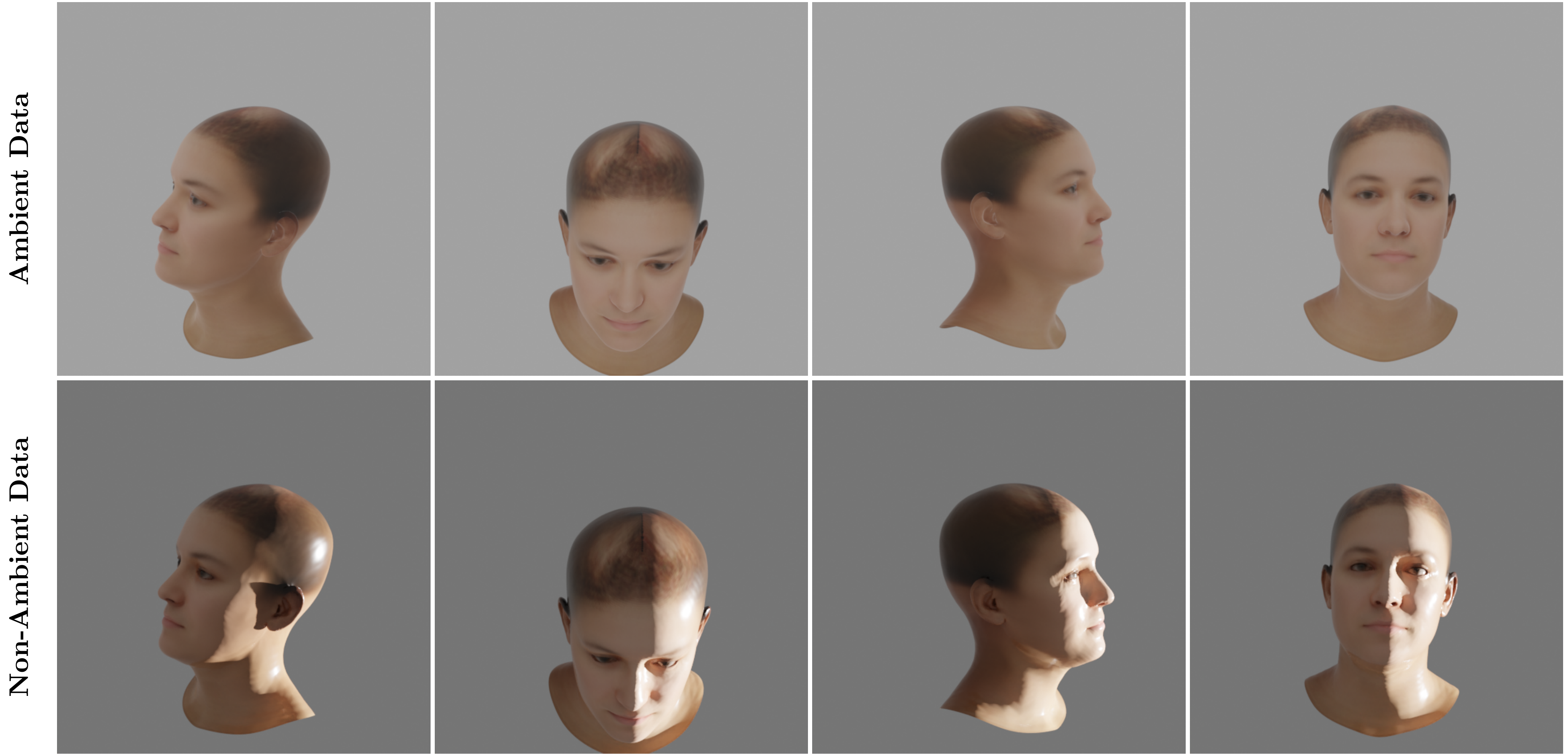}
    \caption{Samples from the Synthetic Datasets described in \sect{syn-data}}
    \label{fig:sample-data}
\end{figure}

We create two synthetic datasets using blender: 1) An `Ambient Data' capture setting, which contains a synthetic human head in ambient lighting and 2) A `Non-Ambient Data' capture setting with a strong directional light source that causes cast shadows and specularities. Both datasets are captured using a fixed camera with changing head-pose but a constant neutral facial expression. Sample images from both the datasets are shown in \fig{sample-data}.  

\subsection{Comparisons against IMAvatar on Synthetic Data}
\begin{figure}[h]
    \centering
    \includegraphics[width=0.9\linewidth]{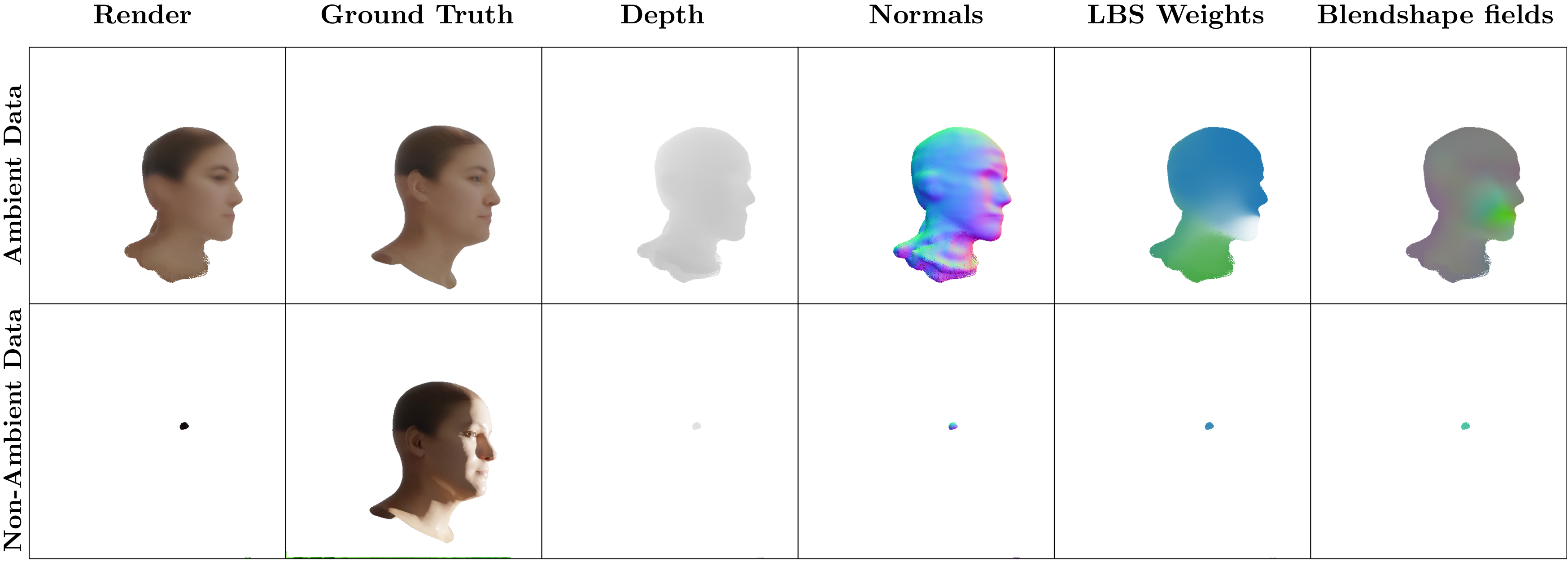}
    \caption{Results of IMAvatar \cite{imavatar} on Synthetic Data. Due to the changing illumination effects induced by head-pose variation in strong non-ambient lighting, IMavatar fails to converge.}
    \label{fig:IMavatar-fail}
\end{figure}

We now compare the results of \MethodName and IMAvatar \cite{imavatar} on synthetic data. Our first finding is that, while IMAvatar \cite{imavatar} converges to a reasonable solution on the Ambient Data, it diverges on Non-Ambient Data (see \fig{IMavatar-fail}). The large appearance changes induced by cast-shadows and specularites in the Non-Ambient data leads to instability in IMavatar's training and it's consequent divergence. We also contacted the authors of IMAvatar \cite{imavatar} regarding this instability; they confirmed that since IMavatar is not designed for such strong non-ambient illumination conditions, training it would be challenging.   Quantitative results on both datasets are shown in \tab{Subjects_metrics_syn_v1}. As can be seen, \MethodName outperfoms IMavatar in both Ambient and Non-Ambient Data settings by a large margin.

\begin{table}[h]
\begin{center}
\small
\scalebox{0.7}{
\begin{tabular}{lccc}
\toprule
  & \multicolumn{1}{c}{\textit{Ambient Data}}  &  \multicolumn{1}{c}{\textit{Non-Ambient Data}} \\
  \midrule
  Models   &  FaceMSE $\downarrow$  &  FaceMSE $\downarrow$\\
\midrule
 \MethodName (Ours) &
  \(\mathcolorbox{pink}{\text{1.03e-4}}\)&
  \(\mathcolorbox{pink}{\text{1.32e-4}}\) &
\\ 
IMAvatar \cite{imavatar} &
  1.56e-2 & 
   - &
 \\ 
\bottomrule
\end{tabular}}
\caption{\small{Quantitative results on synthetic data}
}
\label{tab:Subjects_metrics_syn_v1}
\end{center}
\end{table}

\section{Comparisons against PointAvatar}
\begin{figure*}[h]
    \centering
    \includegraphics[width=\linewidth]{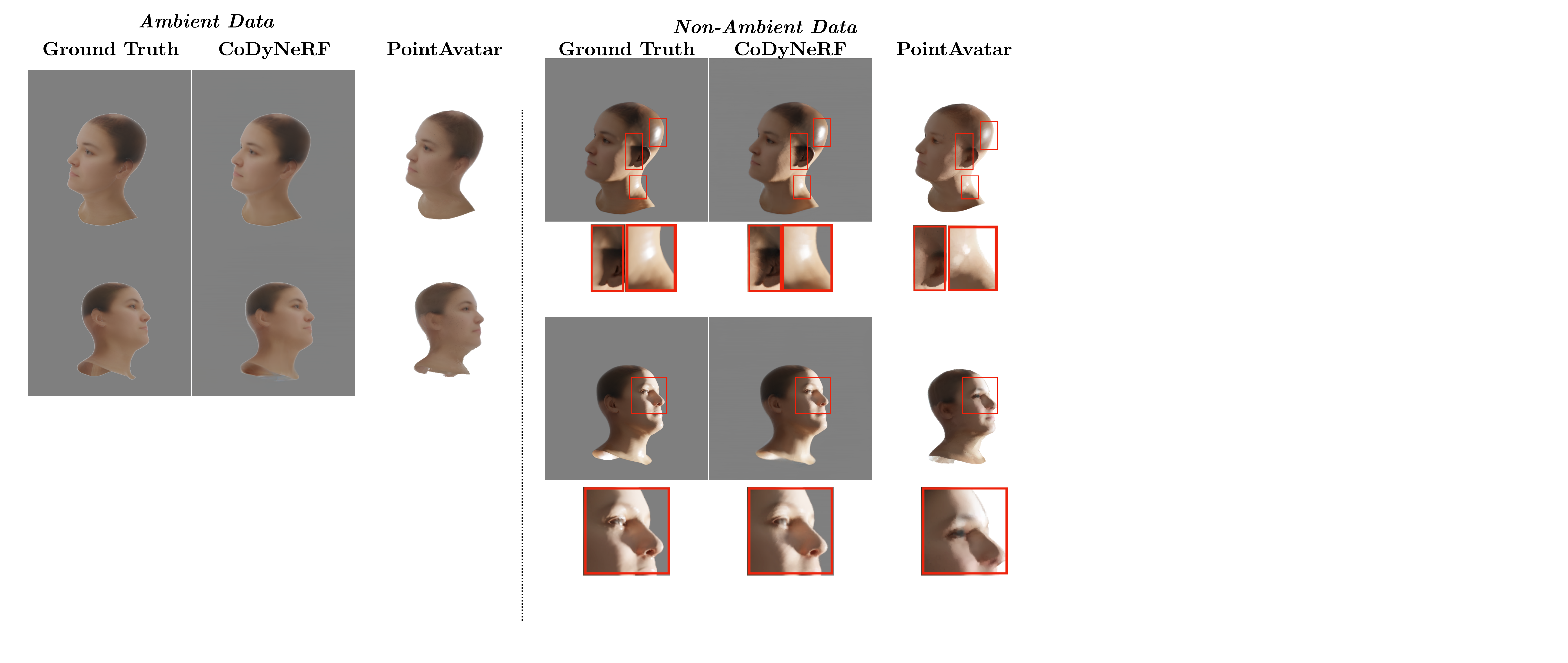}
    \caption{Comparison of \MethodName and PointAvatar \cite{pointavatar} on Synthetic Data. While PointAvatar performs quite well on ambient data (on the Left), it fails to accurate predict illumination effects that are present in non-ambient data (Right). In the top row, we see that PointAvatar is unable to predict the cast shadow of the ear and the specularity on the neck. In the bottom row, we see that PointAvatar \cite{pointavatar} is unable to generate the cast shadow of the nose. In constrast, \MethodName is able to reproduce the cast shadow of the ear, the specularity on the neck (top row) and the cast shadow of nose (bottom row) with high fidelity. }
    \vspace{-0.3cm}
    \label{fig:SynAba_comp_PA}
\end{figure*}

In this section we demonstrate the inablity of Pointavatar \cite{pointavatar} to model specularities and cast shadows using synthetic data described in the \sect{syn-data} \footnote{We use the code from their official repo at: https://github.com/zhengyuf/PointAvatar}. PointAvatar \cite{pointavatar} uses a point cloud and an SDF as the geometric representation of the avatar in the canonical space and the deformation from the canonical space to the deformed space is done by an MLP that predicts deformation blendshapes. The apperance of each point is modelled as a product of an albedo and shading. The per-point albedo is predicted by an MLP that is conditioned on the canonical position and the shading is predicted using an MLP conditioned on the normals. Since the shading network is only condition on the normals, it does not model cast shadows (which depend on global geometry and lighting) and specularities (which depend on both normals and viewing direction), thus it is unable to learn and predict them. As can be seen in \fig{SynAba_comp_PA}, \MethodName and PointAvatar perform comparably on the dataset with ambient lighting conditions. However, when the lighting is non-ambient, \MethodName performs considerably better. For example, in the top row of \fig{SynAba_comp_PA}, we see that PointAvatar \cite{pointavatar} is unable to predict the shadow cast by the ear and the specularity on the neck. Similarly, in the bottom row, PointAvatar \cite{pointavatar} fails to capture the shadow cast by the nose on the face (however, it does predict the shading correctly). In contrast, \MethodName is able to faithfully reproduce the cast shadow of the ear and the specularity on the neck in the top row and the cast shadow of the nose on the face in the bottom row. This can also be seen in the quantitative results in \tab{Subjects_metrics_syn_PA}, where \MethodName outperforms PointAvatar on both datasets, especially on the non-ambient data.

\begin{table}[h]
\begin{center}
\small
\scalebox{0.7}{
\begin{tabular}{lccc}
\toprule
  & \multicolumn{1}{c}{\textit{Ambient Data}}  &  \multicolumn{1}{c}{\textit{Non-Ambient Data}} \\
  \midrule
  Models   &  FaceMSE $\downarrow$  &  FaceMSE $\downarrow$\\
\midrule
 \MethodName (Ours) &
  \(\mathcolorbox{pink}{\text{1.03e-4}}\)&
  \(\mathcolorbox{pink}{\text{1.32e-4}}\) &
\\ 
PointAvatar \cite{pointavatar} &
  5.99e-4 & 
   2.33e-3 &
 \\ 
\bottomrule
\end{tabular}}
\caption{\small{Quantitative results of on Ambient and Non-Ambient Synthetic data described in \sect{syn-data}. We see that \MethodName's results are better than PointAvatar \cite{pointavatar} on both datasets with the difference being larger on the non-ambient dataset due to PointAvatar's inability to predict cast shadows and specularities.}
}

\label{tab:Subjects_metrics_syn_PA}
\end{center}
\end{table}

\begin{table}[t]
\begin{center}
\small
\scalebox{0.75}{
\begin{tabular}{lcccc}
\toprule
  & \multicolumn{1}{c}{\emph{No \(\text{Mesh}_{\mathbf{n}}(\mathbf{x})\) }}  &  \multicolumn{1}{c}{\emph{No \(\text{Grad}_{\mathbf{n}}(\mathbf{x})\)}} &  \multicolumn{1}{c}{\emph{No \(\text{DistToMesh}(\xbf)\)}} &  \multicolumn{1}{c}{\emph{Full Model}}\\
  \midrule
  Models   & Angular Error  $\downarrow$ &Angular Error $\downarrow$ & Angular Error $\downarrow$ & Angular Error $\downarrow$ \\
\midrule
Head Region &
 \(12.72^{o}\) & \(\mathcolorbox{pink}{3.39^{o}}\) & \(4.16^{o}\) & \(\mathcolorbox{yellow}{3.43^{o}}\) 
 \\
Neck Region  &
 \(8.63^{o}\) &  \(6.73^{o}\)  & \(\mathcolorbox{yellow}{5.12^{o}}\)  & \(\mathcolorbox{pink}{4.89^{o}}\)\\
\bottomrule
\end{tabular}}
\vspace{-0.3cm}
\caption{\footnotesize{Ablation of the inputs to the normals prediction network \(N\) using non-ambient Synthetic data.  \(\mathcolorbox{pink}{\text{Lowest Value}}\), \(\mathcolorbox{yellow}{\text{Second Lowest value}}\)}.}
\label{tab:normals_input_aba}

\end{center}
\end{table}
\section{Ablating the design of the normals network}
We design the normals prediction network of \MethodName, \(\mathcal{N}\), in a manner that allows it to exploit local priors from the 3D head mesh and the scene to predict the normals at each point of the dynamic neural portrait. The normals at any point \(\xbf\) is as follows:
\begin{equation}
    \textbf{n} = \mathcal{N}(\text{Mesh}_{\textbf{n}}(\textbf{x}), \text{Grad}_{\textbf{n}}(\xbf), \text{DistToMesh}(\xbf))
\end{equation}

The intuition behind this is that depending on the distance of the point to the mesh (\(\text{DistToMesh}(\xbf)\)) \(\mathcal{N}\) can rely on either the mesh normals from the mesh-vertex closest to \(\xbf\)  (\(\text{Mesh}_{\textbf{n}}(\textbf{x})\)) or the scene normals (\(\text{Grad}_{\textbf{n}}(\xbf)\)) to predict the normals. To test the efficacy of this design we ablate each input on non-ambient synthetic data described in \sect{syn-data} by calculating the angular error between the predicted and ground-truth normals on vertices from the head and neck regions. While we have sparse supervision on 5000 vertices of the head region region through \(\mathcal{R}_{Mesh, \textbf{n}}\)  (Eq. 9 of the paper), we do not have any supervision on the neck region.  We measure the angular error on both the head and neck regions of an upsampled version of the mesh, sampling 10k \textit{novel} vertices from each region.   \tab{normals_input_aba} shows that the model with all three inputs \(\textbf{Mesh}_{\textbf{n}}(\xbf), \textbf{Grad}_{\textbf{n}}(\xbf)\) and \(\text{DistToMesh}(\xbf)\) (called `Full Model") has a low angular error on both head and neck regions.  Not having \(\textbf{Mesh}_{\textbf{n}}(\xbf)\) as input hurts normals prediction accuracy  of both regions, not having \(\textbf{Grad}_{\textbf{n}}(\xbf)\) hurts the the neck region accuracy only.

\begin{table*}[h]
\centering
\begin{tabular}{@{}llll@{}}
\toprule
Method                     & PSNR  (\(\uparrow\))     & LPIPS  (\(\downarrow\))  & FaceMSE  (\(\downarrow\)) \\ \midrule
\MethodName & \hl{23.46} & \hl{0.40} & \hl{1.93e-3} \\
\MethodName wo Cauchy Regularization & 22.5 & 0.51 & 2.34e-3  \\

\end{tabular}
\caption{Ablation for the Cauchy Regularization.}
\label{tab:aba}
\end{table*}

\section{Details About the Testing Set}
\begin{figure}[h]
    \centering
    \includegraphics[width=0.95\linewidth]{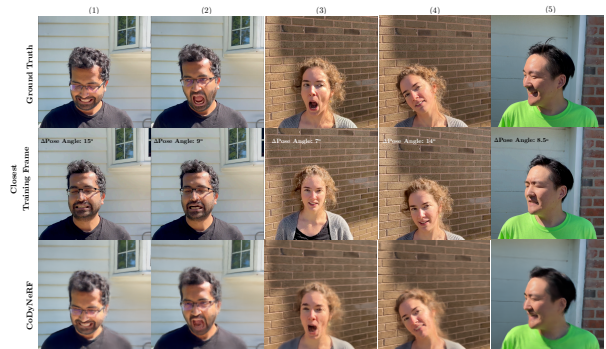}
    \vspace{-0.3cm}
    \caption{Test set reanimation with closest training frames}
    \label{fig:testset}
    \vspace{-0.5cm}
\end{figure}

We make sure that the test set is completely held out from the training set by using the last 120-150 frames of the video as the test set. Similar to RigNeRF [3], the test set contains head-poses that have relative angles to the closest training ranging from \(0^{o}-15^{o}\). In \fig{testset}, we have recreated Fig 5 from the paper with the closest training set frame to provide context to the reanimation results.

\section{Ablation of Spatial conditioning of the density prediction model}
\MethodName predicts the density at each point, \(\xbf\), using the canonical position of the point \(\xbf_{can}\) and the distance of the point from the mesh in the deformed space i.e \(\text{DistToMesh}(\xbf)\).  Such a conditioning allows \(F\) to distinguish between points in the canonical space that have never been deformed and points that have been deformed to the canonical space, leading to faster convergence and better results. In \tab{spatial-density-aba}, we measure the test-set PSNR of \MethodName with and without conditioning on \(\text{DistToMesh}(\xbf)\) (\MethodName and \MethodName (w/o spatial cond.) respectively) on two subjects. As can be seen, \MethodName converges to a higher PSNR much faster when the density is conditioned on the distance to the mesh (i.e \(\text{DistToMesh}(\xbf)\)).
\begin{table*}[h]
\centering
\begin{tabular}{@{}lllll@{}}
\toprule
Subject                     &Method                     & Epochs Trained       & PSNR  & PSNR after 25k more training epochs  \\ \midrule
Subject 1 &\MethodName &200000 & \(\mathcolorbox{pink}{\text{23.46}}\) & \(\mathcolorbox{pink}{\text{23.51}}\)  \\
&\MethodName (w/o spatial cond.)                   & 200000  & 22.31 & 22.37  \\ 
 \bottomrule
Subject 2 &\MethodName &200000 & \(\mathcolorbox{pink}{\text{21.24}}\)  & \(\mathcolorbox{pink}{\text{21.31}}\)  \\
&\MethodName (w/o spatial cond.)                    & 200000  & 20.97 & 21.13 \\ 
\bottomrule
\end{tabular}
\caption{Training convergence of \MethodName with and without spatial conditioning of the density using \(\text{DistToMesh}(\xbf)\). As can be seen, using spatial conditioning boosts both the rate of convergence and the final PSNR.}
\label{tab:spatial-density-aba}
\end{table*}

\section{Ablation of Cauchy Regularization} We ablate the use of the cauchy regularization (Eq. 14 of the paper) to improve underlying scene normals. We use cauchy regularization to ensure the density distribution along the ray is sparse, thus ensuring there are only a few points of high density along the ray. This allows the importance sampling based regularization of the normals (Eq 13) to work more effectively, as it can act only on the top \(k = 30\) points and can ignore the remaining points. 

\begin{figure}[h]
    \centering
    \includegraphics[width=\linewidth]{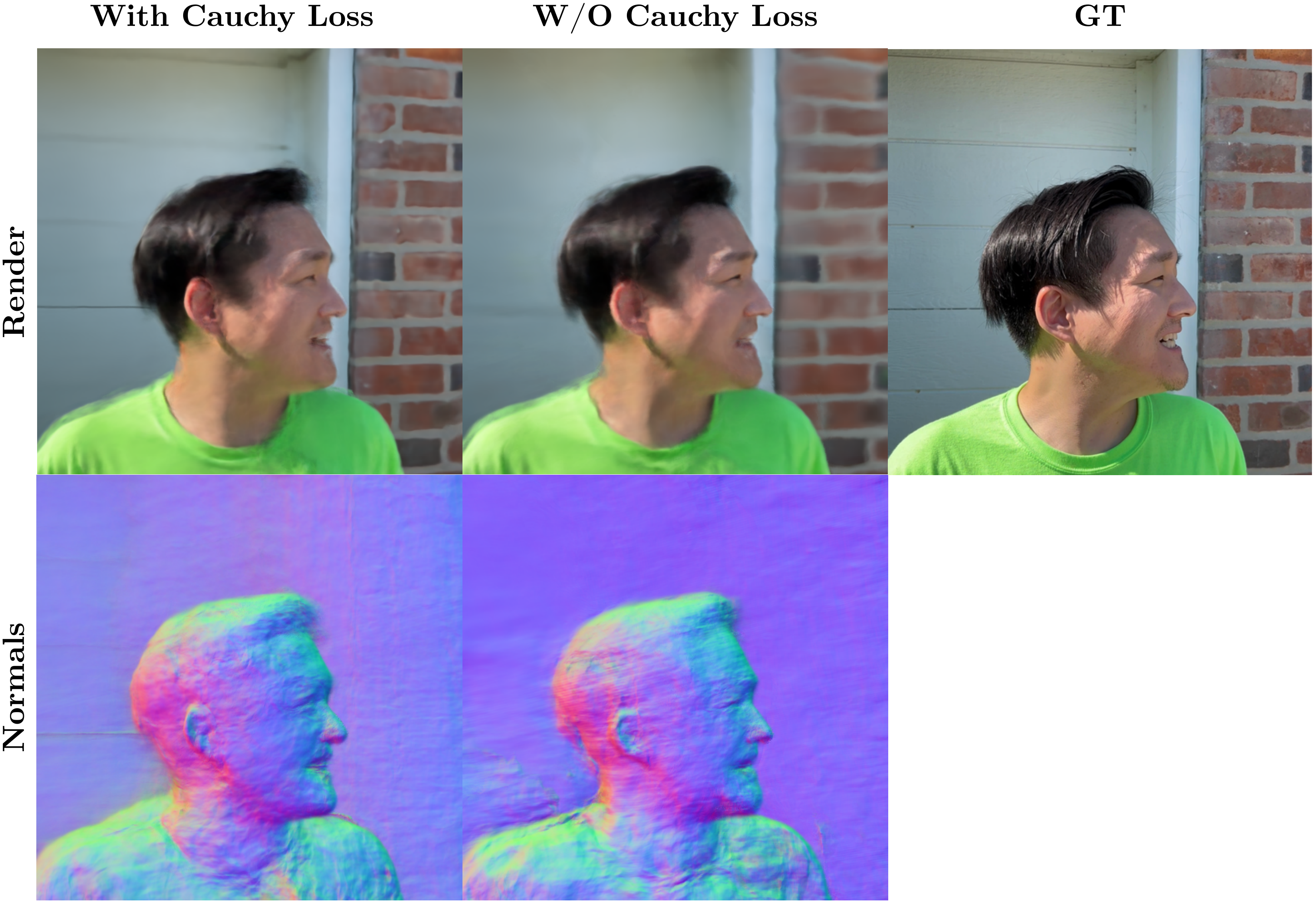}
    
    \caption{{\textbf{Qualitative comparison between model with and without a Cauchy Loss.} 
    }}
    \label{fig:aba-qual}
\end{figure}

As can be seen quantitatively in \tab{aba} and qualitatively in \fig{aba-qual}, the cauchy regularization leads to significantly sharper renders and fidelity of normals.  In the top row of \fig{aba-qual}, we observe that both the render of both the face region and the background is sharper when using the Cauchy loss. In the bottom row, we see that the face and background normals are both significantly more accurate. This improvement in quality is also reflected in the quantitative metrics as demonstrated in  \tab{aba}.

\section{Additional Metrics}
\begin{figure}[h]
    \centering
    \includegraphics[width=0.1\linewidth]{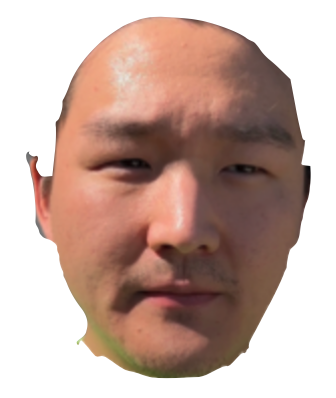}
    
    \caption{{An example of the region over which the FaceMSE is calculated. 
    }}
    \label{fig:facemse-region}
\end{figure}

.
We also evaluate the SSIM metric on the real data and report it in \tab{Subjects_metrics_ssim}. As can be seen, \MethodName outperforms prior work across all subjects. In \fig{facemse-region}, we show an illustrative example of the region over which the FaceMSE i.e MSE over the face region, is calculated.
\begin{table}[t]
\begin{center}
\small
\scalebox{0.80}{
\begin{tabular}{lcccccc}
\toprule
  & \multicolumn{1}{c}{\emph{Subject 1}}  &  \multicolumn{1}{c}{\emph{Subject 2}} &  \multicolumn{1}{c}{\emph{Subject 3}} &  \multicolumn{1}{c}{\emph{Subject 4}}\\
  \midrule
  Models   & SSIM $\uparrow$ &SSIM $\uparrow$ & SSIM $\uparrow$ & SSIM $\uparrow$ \\
\midrule
 \MethodName (Ours) &
 \mathcolorbox{pink}{0.83} & \mathcolorbox{pink}{0.76} & \mathcolorbox{pink}{0.73} & \mathcolorbox{pink}{0.81} 
 \\
 RigNeRF [3] &
 0.71 &  0.73  & 0.68  & 
 0.76
  \\
 NHA \cite{nha} &
  0.58 &  0.61 & 0.55  & 0.48
\\ 
PointAvatar \cite{imavatar} &
  0.64 &  0.65  & 0.67  & 
  0.73
 \\ 
\bottomrule
\end{tabular}}
\vspace{-0.3cm}
\caption{\footnotesize{SSIM on test data. \(\mathcolorbox{pink}{\text{Highest Value}}\).}
}

\label{tab:Subjects_metrics_ssim}
\vspace{-0.9cm}
\end{center}
\end{table}

\section{Experimental Details}

All our models were trained on 4 A100 GPUs (with 40GB VRAM each) using 128 samples per ray and a batch-size of 1000 rays. For all methods, we use 10 frequencies to encode the position of a point and 4 frequencies to encode the direction of the ray as  input to \(F\). For \MethodName and RigNeRF,  we use 10 frequencies to encode the position of a point as input to \(D\). We also use a coarse-to-fine regularization on the positional encoding of the position input to \(D\) for 10k epochs. The architecture of \(F\) is identical to the original NeRF MLP \cite{nerf} with 8 layers of 256 hidden units each. We use the same \(F\) for all our methods. \(D\) consists of 8 layers with 128 hidden units in each. We use the same architecture for \(D\) for both \MethodName and RigNeRF. The normals network \(\mathcal{N}\) consists of three layers with 128 hidden units. We train \MethodName and RigNeRF with an initial learning rate of \(5e^{-4}\) which is decayed to \(5e^{-5}\) by the end of training. For Neural Head Avatars \cite{nha} we follow the training settings given in the \href{https://github.com/philgras/neural-head-avatars/blob/main/configs/optimize_avatar.ini}{config file of the publically available github repo.} Similarly, for PointAvatar we follow the training setting given in the \href{ https://github.com/zhengyuf/PointAvatar}{ publically available github repo}, with the only difference being that we train with 220k points instead of 409000 points due to GPU memory constraints.

\begin{table*}[h]
\centering
\begin{tabular}{@{}lllll@{}}
\toprule
Subject                     &Method                     & Epochs Trained       & App Code dim  & Def Code dim  \\ \midrule
Subject 1 &\MethodName &200000 & 8 & 8 \\
&RigNeRF                   & 200000  & 8 & 8 \\ 
 \bottomrule
Subject 2 &\MethodName &200000 & 8 & 8 \\
&RigNeRF                   & 200000  & 8 & 8 \\ 
\bottomrule
Subject 3 &\MethodName &200000 & 8 & 8 \\
&RigNeRF                   & 200000  & 8 & 8 \\ 
 \bottomrule
Subject 4 &\MethodName &200000 & 8 & 8 \\
&RigNeRF                   & 200000  & 8 & 8 \\ 
 \bottomrule
\end{tabular}
\caption{Training configuration for all the experiments.}
\label{tab:model-details}
\end{table*}

\section{Societal Impact}
Since \MethodName is capable of reanimating face, it is prone to misuse by bad actors to generate deep-fakes. However, the authors of \cite{wang2019cnngenerated} show that it is possible to train discriminative classifiers to detect images and videos generated by synthetic methods like \MethodName. Another possible solution to detect synthetically generated faces is to use works such as \cite{yu2021artificial} to watermark training images in order to detect real or fake images.

\end{document}